\documentclass[11pt,a4paper]{article}
\usepackage[hyperref]{acl2018}
\usepackage{times}
\usepackage{latexsym}
\usepackage{helvet}  %
\usepackage{courier}  %
\usepackage{url}
\usepackage{amsmath}
\usepackage{amssymb}
\usepackage{amsfonts}
\usepackage{graphicx}
\usepackage{subcaption}
\usepackage{color}
\usepackage{xspace}

\usepackage{booktabs}
\usepackage{adjustbox}

\newcommand{\cut}[1]{}
\definecolor{ultraviolet}{RGB}{95, 75, 139}
\definecolor{darkorchid}{RGB}{153 , 50 ,204}
\definecolor{seagreen}{RGB}{46 ,139,  87}

\graphicspath{ {FIG/} }
\newcommand{\xhdr}[1]{{\noindent\bfseries #1.}}
\newcommand{\xhdrc}[1]{{\noindent\bfseries #1}}

\newcommand{\awryturning}{awry-turning\xspace}
\newcommand{\ontrack}{on-track\xspace}

\newcommand{\healthy}{civil\xspace}

\aclfinalcopy
\def\aclpaperid{1630}
\title{Conversations Gone Awry: \\
Detecting Early Signs of Conversational Failure}

\author{
	Justine Zhang \and Jonathan P. Chang \and Cristian Danescu-Niculescu-Mizil\thanks{\ \ \ Corresponding senior author.}\\
	Cornell University\\
	{\tt \{jz727,jpc362\}@cornell.edu, cristian@cs.cornell.edu} \\\AND
  {\ \ \  Lucas Dixon \and Nithum Thain}\\
  Jigsaw \\
  {\tt \hspace{-0.5cm} \{ldixon,nthain\}@google.com} \\\And
  {\ \ \ \ \ \ \  Yiqing Hua} \\
  {\ \ \ \ \ \ \ Cornell University} \\
  {\tt \ \ \ yh663@cornell.edu} \\\And
  \ \ \ Dario Taraborelli \\
  \ \ \ Wikimedia Foundation \\
  {\tt \hspace{0.25cm} dario@wikimedia.org} \\
}

\begin{document}

\maketitle

\begin{abstract}

One of the main challenges online social systems face is the prevalence of antisocial behavior, such as harassment and personal attacks.  In this work, we introduce the task of predicting from the very start of a conversation whether it will get out of hand.  As opposed to detecting undesirable behavior after the fact, this task aims to enable early, actionable prediction at a time when the conversation might still be salvaged.

To this end, we develop a framework for capturing pragmatic devices---such as politeness strategies and rhetorical prompts---used to start a conversation, and analyze their relation to  its future trajectory. 
Applying this framework in a controlled setting, we demonstrate the feasibility of detecting early warning signs of antisocial behavior in online discussions. 

\end{abstract}

\section{Introduction}
\label{sec:intro}

\begin{quote}
\small
\vspace{-0.01cm}
``Or vedi l'anime di color cui vinse l'ira.''\footnote{``Now you see the souls of those whom anger overcame.''}
\end{quote}
\vspace{-0.34cm}
\begin{flushright}
\small
-- Dante Alighieri, Divina Commedia, Inferno
\end{flushright}
Online conversations have a reputation for going awry~\cite{Hinds2005-rt,Gheitasy2015-na}:
antisocial behavior~\cite{Shepherd2015-yd} or simple misunderstandings~\cite{Churchill2000-sg,Yamashita2006-eh} hamper the efforts of even the best intentioned collaborators.
Prior computational work has focused on characterizing and detecting content exhibiting antisocial online behavior: trolling~\cite{Cheng2015-xc,Cheng2017-bw}, hate speech~\cite{warner2012detecting,Davidson2017-tu}, harassment~\cite{Yin2009-ne}, personal attacks~\cite{Wulczyn2017-vg} or, more generally, toxicity~\cite{Chandrasekharan:2017:YCS:3171581.3134666,Pavlopoulos2017-jt}. 

Our goal is crucially different: instead of identifying antisocial comments {\em after the fact}, we aim to detect {\em warning signs} indicating that a \healthy conversation is at risk of derailing into 
such
 undesirable behaviors.  
Such warning signs could provide potentially actionable knowledge at a time when the conversation is still salvageable.
\newcommand{\examplespace}{\vspace{0.1cm}}

\begin{figure*}
\centering
\begin{adjustbox}{width=2.07\columnwidth}
\begin{tabular}{p{10cm} p{10cm}}
  \toprule
  \begin{minipage}[t]{1.25\columnwidth}
	  \textbf{A1:}  Why there's no mention of it here? Namely, an altercation with a foreign intelligence group? True, by the standards of sources some require it wouln't even come close, not to mention having some really weak points, but it doesn't mean that it doesn't exist.
	  \vspace{0.1in}

	  \hspace{0.5in}\begin{minipage}[t]{0.85\columnwidth}
		  \textbf{A2:} So what you're saying is we should put a bad source in the article because it exists?
	  \end{minipage}
  \end{minipage} & \begin{minipage}[t]{1.25\columnwidth}
	  \textbf{B1:} Is the St.~Petersberg Times considered a reliable source by wikipedia? It seems that the bulk of this article is coming from that one article, which speculates about missile launches and UFOs. I'm going to go through and try and find corroborating sources and maybe do a rewrite of the article. I don't think this article should rely on one so-so source.
	  \vspace{0.1in}

	  \hspace{0.5in}\begin{minipage}[t]{0.85\columnwidth}
		  \textbf{B2:} I would assume that it's as reliable as any other mainstream news source. 
	  \end{minipage}
              \vspace{0.05in}
  \end{minipage} \\
\bottomrule
\end{tabular}
\end{adjustbox}
\caption{Two examples of initial exchanges from conversations concerning disagreements between editors working on the Wikipedia article about the Dyatlov Pass Incident. Only one of the conversations will eventually turn awry, with an interlocutor launching into a personal attack.}
  \label{fig:intro_examples}
  \vspace{-0.05in}
\end{figure*}

As a motivating example, consider the  pair of 
conversations
 in Figure \ref{fig:intro_examples}. Both exchanges took place in the context of the Wikipedia discussion page for the article on the Dyatlov Pass Incident,
  and both 
show
  (ostensibly) civil disagreement between the participants.  However, only one of these conversations will eventually 
  turn awry
  and devolve into a personal attack (``Wow, you're coming off as a total d**k. [...]  What the hell is wrong with you?''), while the other will remain civil.

As humans, we have some intuition about which conversation is more likely to derail.\footnote{In fact, humans achieve an accuracy of 72\% on this balanced task, showing that it is feasible, but far from trivial.}
We may note the repeated, direct questioning with which \textbf{A1} opens the exchange, and that \textbf{A2} replies with yet another question.  
In contrast, \textbf{B1}'s softer, hedged approach (``it seems'', ``I don't think'') appears to invite
an exchange of ideas, and
\textbf{B2} actually addresses the question 
 instead of stonewalling.   Could we endow artificial systems with such intuitions about the future trajectory of conversations?

In this work we aim to computationally capture linguistic cues that predict a conversation's future health.  Most existing 
conversation modeling approaches
aim to detect 
characteristics of an observed discussion or predict the outcome after the discussion concludes---e.g., whether it involves a present dispute~\cite{Allen2014-xa,wang2016piece} or  contributes to the eventual solution of a 
problem~\cite{Niculae2016-ay}.
 In contrast, for this 
 new
  task we need to discover 
  interactional 
signals of the {\em future} trajectory of an {\em ongoing} conversation.

We make a first approach to this problem by analyzing the role of politeness (or lack thereof) in keeping conversations on track. Prior work has shown that politeness can help shape the course of offline \cite{clark1979responding,clark1980polite}, as well as online interactions \cite{Burke:2008:MYP:1460563.1460609}, through mechanisms such as softening the perceived force of a message \cite{fraser1980conversational}, acting as a buffer between conflicting interlocutor goals \cite{brown1987politeness}, and enabling all parties to save face \cite{goffman1955face}. This suggests the potential of politeness to serve as an indicator of whether a conversation will sustain its initial civility or eventually derail, and motivates its 
consideration
 in the present work.

Recent 
studies have
  computationally operationalized prior formulations of politeness by extracting linguistic cues that reflect politeness strategies \cite{Danescu-Niculescu-Mizil2013-so, aubakirova2016interpreting}. Such research has additionally tied politeness to social factors such as individual status \cite{danescu2012echoes, krishnan2014you}, and the success of requests \cite{althoff2014ask} or of collaborative projects \cite{ortu2015bullies}. However, to the best of our knowledge, this is the first computational investigation of the 
relation between
  politeness strategies and the future trajectory of the conversations in which they are deployed. 
     Furthermore,
  we generalize beyond
   predefined politeness strategies by using an unsupervised method to  discover additional rhetorical 
   prompts
    used to initiate different types of 
   conversations
    that may be specific to online collaborative settings, such as 
    coordinating work \cite{Kittur2008-bc}
     or 
   conducting factual checks. 

We explore the role of such pragmatic and rhetorical devices in foretelling a particularly perplexing type of conversational failure: 
when participants 
engaged in previously civil discussion start to attack each other.  This type of derailment ``from within'' is arguably more disruptive than other forms of antisocial behavior, such as vandalism or trolling, which the interlocutors have less control over or can choose to ignore.

We study this phenomenon
in a new dataset of Wikipedia talk page discussions, which we compile through a combination of machine learning and crowdsourced filtering. The dataset consists of conversations which begin with ostensibly civil comments, 
and either remain
healthy or 
derail into personal attacks.
Starting from this data,
 we construct a setting that mitigates effects which may trivialize the task. In particular, some topical contexts 
(such as politics and religion)
 are naturally more susceptible to antisocial behavior \cite{kittur_whats_2009,Cheng2015-xc}.
  We employ techniques from causal inference \cite{Rosenbaum2010-jg} to establish a controlled framework that focuses our study on topic-agnostic linguistic cues.

In this controlled 
setting, we find that pragmatic cues extracted from the very first exchange in a conversation 
(i.e., 
the first
comment-reply pair) can indeed provide some signal of whether the conversation will
subsequently go awry. 
For example, conversations prompted by hedged
 remarks 
 sustain their initial civility more so than those prompted by forceful questions, or by direct language addressing the other interlocutor.

In summary, our main contributions are:
\begin{itemize}
	\item We articulate the new task of detecting early on whether a conversation will derail into personal attacks;
	\item We devise a controlled setting and build 
a labeled dataset to study this phenomenon;
\newpage
	\item We investigate how politeness strategies and other 
   rhetorical
   devices 
   are tied to the future trajectory of a conversation.
\end{itemize}

More broadly, we show the feasibility of automatically detecting warning signs of future misbehavior in 
collaborative interactions.
By providing a labeled dataset together with basic methodology and several baselines,
we open 
the door to further work on understanding factors which may derail or sustain healthy online conversations.  To facilitate such future explorations, we distrubute the data and code as part of the Cornell Conversational Analysis Toolkit.\footnote{\url{http://convokit.infosci.cornell.edu}}

\section{Further Related Work}
\label{sec:related}

\xhdr{Antisocial behavior}
Prior work has studied a wide range of disruptive interactions in various online platforms like Reddit and Wikipedia, examining behaviors like aggression \cite{kayany1998contexts}, harassment \cite{chatzakou2017measuring, vitak2017identifying}, and bullying \cite{akbulut2010cyberbullying, kwak2015exploring, singh2017they}, as well as their impact on aspects of engagement like user retention \cite{Collier12,wikisurvey} or discussion quality \cite{arazy2013stay}. Several studies have sought to develop machine learning techniques to detect signatures of online toxicity, such as personal insults~\cite{Yin2009-ne}, harassment~\cite{sood2012automatic} and abusive language \cite{nobata2016abusive, gamback2017cnnhate, pavlopoulos2017deepmoderation, Wulczyn2017-vg}. These works focus on detecting toxic behavior after it has already occurred; a notable exception is \newcite{Cheng2017-bw}, which 
predicts
 future community enforcement against users in news-based discussions. 
Our work similarly aims to understand
\textit{future} antisocial behavior; however, our focus is on studying the trajectory of a conversation rather than the behavior of individuals across disparate discussions.  

\xhdr{Discourse analysis}
Our present study builds on a large body of prior work in computationally modeling discourse. Both unsupervised \cite{ritter2010unsupervised} and supervised \cite{zhang2017characterizing} approaches have been used to categorize behavioral patterns on the basis of the language that ensues in a conversation, in the particular realm of online discussions.
Models of conversational behavior have also been 
used 
to predict conversation outcomes, such as 
betrayal in games \cite{Niculae2015-hj}, and success in team problem solving settings \cite{fu2017confidence} or in persuading others \cite{tan2016winning,zhang2016conversational}.   

While we are inspired by the techniques employed in these approaches, our work is concerned with predicting the future trajectory of an ongoing conversation as opposed to a post-hoc outcome. In this sense, we build on prior work in modeling conversation trajectory, which has largely considered \textit{structural} aspects of the conversation \cite{Kumar:2010:DC:1835804.1835875,Backstrom2013-oe}. We complement these structural models by seeking to 
 extract potential signals of future outcomes from the \textit{linguistic discourse} within the conversation.

\section{Finding Conversations Gone Awry}
\label{sec:data}
We develop our framework for understanding linguistic markers of conversational trajectories in the context of Wikipedia's \textit{talk page} discussions---public forums in which contributors convene to deliberate on editing matters such as evaluating the quality of an article and reviewing the compliance of contributions with community guidelines.
The dynamic of conversational derailment is particularly intriguing and consequential in this setting by virtue of its collaborative, goal-oriented nature. In contrast to 
unstructured commenting forums, cases where one \textit{collaborator} turns on another over the course of an initially civil exchange constitute perplexing pathologies. In turn, these toxic attacks are especially 
disruptive
 in Wikipedia since they undermine the social fabric of the community as well as the ability of editors to contribute~\cite{wikiboard2016}. 

\begin{table*}[ht]
\centering
\begin{adjustbox}{width=2.07\columnwidth}
\begin{tabular}{p{10cm} p{10cm}}
  \toprule
  \begin{minipage}[t]{1.25\columnwidth}
    \xhdrc{Job 1: Ends in personal attack.} We show 
    three
     annotators a conversation and ask them to determine if its {last} comment is a personal attack toward someone else in the conversation.\\

    \hspace{0.5in}\begin{minipage}[t]{1\columnwidth}
\begin{tabular}{cccc}

    \textbf{Annotators} &  \textbf{Conversations} & \textbf{Agreement} \\
 367 & 4,022 & 67.8\% \\
\end{tabular}
    \end{minipage}
  \end{minipage} & \begin{minipage}[t]{1.25\columnwidth}
    \xhdrc{Job 2: Civil start.} We split conversations into snippets of three consecutive comments. We ask three annotators to determine whether any of the comments in a snippet is toxic.\\
    
    \hspace{0.15in}\begin{minipage}[t]{1\columnwidth}
\begin{tabular}{cccc}

    \textbf{Annotators} &  \textbf{Conversations} &  \textbf{Snippets}& \textbf{Agreement} \\
247 & 1,252 & 2,181 & 87.5\%\\
\end{tabular}
    \end{minipage}

  \end{minipage} \\
\bottomrule
\end{tabular}
\end{adjustbox}
\caption{Descriptions of crowdsourcing jobs, with relevant statistics. More details in Appendix~\ref{sec:appendix_annot}.}
  \label{tab:annotation}
\end{table*}

To approach this domain we reconstruct a complete view of the conversational process in the edit history of English Wikipedia by translating sequences of revisions of each talk page into structured conversations.  This yields roughly 50 million conversations across 16 million talk pages.

Roughly one percent of Wikipedia comments are estimated to exhibit antisocial behavior~\cite{Wulczyn2017-vg}.
This illustrates a challenge for studying conversational failure:
one has to sift through many conversations in order to find even a small set of examples.
To avoid such a prohibitively exhaustive analysis, we first use a machine learning
classifier to identify candidate conversations that are likely to contain a toxic contribution,
and then use crowdsourcing to vet the resulting labels and construct our controlled dataset.  
\xhdr{Candidate selection} Our goal is to analyze how the start of a \emph{civil} conversation is tied to its potential future derailment into personal attacks.  Thus, we only consider conversations that start out as ostensibly civil, i.e., where at least the first exchange does not exhibit any toxic behavior,\footnote{For the sake of generality, in this work we focus on this most basic conversational unit: the first comment-reply pair starting a conversation.} 
and that continue beyond this first exchange.
To focus on the especially perplexing cases when the attacks come \emph{from within}, we seek examples where the attack is initiated by one of the two participants 
in
 the initial exchange.  

To select candidate conversations to include in our collection, we use the toxicity classifier provided by the  Perspective API,\footnote{\url{https://www.perspectiveapi.com/}} which is trained on Wikipedia talk page comments that have been annotated by crowdworkers~\cite{toxicityDataset2017}. 
This provides a toxicity score $t$ for all comments in our dataset, which we use to preselect two 
sets of conversations: 
(a)~candidate 
conversations that are civil throughout, i.e., conversations in which all comments (including the initial exchange) are not labeled as toxic ($t<0.4$); and (b)~candidate 
conversations that turn toxic after the first (civil) exchange, i.e., conversations in which the $N$-th comment ($N>2$) is labeled toxic ($t\geq0.6$), but all the preceding comments are not ($t<0.4$).

\xhdr{Crowdsourced filtering} Starting from these candidate 
sets, we use crowdsourcing to vet 
each conversation and select a subset that are perceived by humans to either stay civil throughout (``on-track'' conversations), or start civil but end with a \emph{personal attack} (``awry-turning'' conversations).
To inform the design of this human-filtering process and to 
check its effectiveness, we start from a seed set of 232 conversations manually verified by the authors to end in personal attacks (more details about the selection of the seed set and its role in the crowd-sourcing process can be found in Appendix~\ref{sec:appendix_annot}).
We take particular care to not over-constrain crowdworker interpretations of what personal attacks may be, and to separate toxicity from civil disagreement, which is recognized as a key aspect of effective collaborations~\cite{Coser1956-lv,De_Dreu2003-bb}.
We design and deploy two filtering jobs using the \mbox{CrowdFlower} platform, 
summarized
 in Table~\ref{tab:annotation} and detailed in Appendix~\ref{sec:appendix_annot}. \textbf{Job 1} is designed to select conversations that contain a ``rude, insulting, or disrespectful'' comment towards another user in the conversation---i.e., a personal attack.  
In contrast to prior work labeling antisocial comments in isolation \cite{sood2012automatic,Wulczyn2017-vg}, annotators are asked to label personal attacks in the \textit{context} of the conversations in which they occur, 
since antisocial behavior can often be context-dependent \cite{Cheng2017-bw}.  
In fact, in order to ensure that the crowdworkers read the entire conversation, we also ask them to indicate who is the target of the attack. We apply this task to the set of candidate awry-turning conversations, selecting the 14\% which all three annotators perceived as ending in a personal attack.\footnote{We opted to use unanimity in this task to account for the highly subjective nature of the phenomenon.}

\textbf{Job 2} is designed to filter out conversations that do not actually start out as 
civil.  We run this job to ensure that the \textit{awry-turning} conversations are civil up to the point of the attack---i.e., they \emph{turn} awry---discarding 5\% of the candidates that passed Job 1.  We also use it to verify that the candidate \textit{on-track} conversations are indeed civil throughout, discarding 1\% of the respective candidates.  In both cases we filter out conversations in which three annotators could identify at least one comment that is ``rude, insulting, or disrespectful''.
\label{sec:method}
\xhdr{Controlled setting} 
Finally, we need to construct a setting that affords for meaningful comparison between conversations that derail and those that stay on track, and that accounts for trivial topical confounds \cite{kittur_whats_2009,Cheng2015-xc}.  
We mitigate topical confounds using matching, a technique developed for causal inference in observational studies \cite{Rubin2007-ck}.
Specifically, starting from our 
 human-vetted collection of conversations, we pair each \textit{\awryturning} conversation, with an \textit{\ontrack} conversation, such that both took place on the same talk page. 
If we find multiple such pairs, we only keep the one in which the paired conversations take place closest in time, to tighten the control for topic.  
Conversations that cannot be paired are discarded.  

This procedure yields a total of 1,270 paired \awryturning and \ontrack conversations 
(including our initial seed set), spanning 582 distinct talk pages (averaging 1.1 pairs per page, maximum 8) and 1,876 (overlapping) topical 
categories.  The average length of a conversation is 4.6 comments.

\section{Capturing Pragmatic Devices}
\label{sec:features}

\begin{table*}[t]
\centering
\begin{adjustbox}{width=2.07\columnwidth}
  \begin{tabular}{lll}
    \toprule
    Prompt Type & Description & Examples \\
    \midrule
    \texttt{Factual check} & Statements about article content, pertaining to or & The terms \textbf{are used} interchangeably in the US. \\
    & contending issues like factual accuracy. &  The census \textbf{is not talking about} families here. \\
    \addlinespace[5pt]
    \texttt{Moderation} & Rebukes or disputes concerning moderation decisions & \textbf{If} you continue, you may \textbf{be blocked} from editing. \\
    & such as blocks and reversions. & He's \textbf{accused} me \textbf{of} being a troll. \\
    \addlinespace[5pt]
    \texttt{Coordination}  & Requests, questions, and statements of intent  & It's a long list so I \textbf{could do with} your \textbf{help}. \\
    & pertaining to collaboratively editing an article. & \textbf{Let me know} if you agree with this and I'll go ahead [...] \\
    \addlinespace[5pt]
    \texttt{Casual remark} & Casual, highly conversational aside-remarks. & \textbf{What's with} this flag image? \\
    & & I\textbf{'m surprised} there wasn't an article before. \\
    \addlinespace[5pt]
    \texttt{Action statement}  & Requests, statements, and explanations about  & \textbf{Please consider improving} the article to address the issues [...] \\
    & various editing actions. & The page \textbf{was deleted as} self-promotion. \\
    \addlinespace[5pt]
    \texttt{Opinion}  & Statements seeking or expressing opinions about  & I \textbf{think} that it \textbf{should be} the other way around. \\
    & editing challenges and decisions. & This article \textbf{seems to have} a lot of bias. \\
    \bottomrule
  \end{tabular}
\end{adjustbox}
\caption{Prompt types automatically extracted from talk page conversations, with interpretations and examples from the data. Bolded text indicate common prompt phrasings extracted by the framework. Further examples are shown in Appendix~\ref{sec:prompt_type_egs}, Table~\ref{tab:prompt_types_appendix}.}
\label{tab:prompt_types}
\vspace{0.09in}
\end{table*}

We now describe our framework for capturing 
linguistic cues
 that might inform a conversation's future trajectory. Crucially, given 
 our focus on conversations that  start seemingly
civil, we do not expect overtly hostile 
language---such as insults \cite{Yin2009-ne}---to
 be informative.  Instead, we seek to identify pragmatic 
markers
  within the initial exchange of a conversation that might serve to reveal or exacerbate underlying tensions that eventually come to the fore, or conversely suggest sustainable civility.  In particular, in this work we explore how politeness strategies and rhetorical prompts reflect the future health of a conversation.

\xhdr{Politeness strategies}
Politeness can reflect \mbox{a-priori} good will and help navigate potentially  face-threatening acts \cite{goffman1955face,lakoff1973logic}, and also offers hints to the underlying intentions of the interlocutors \cite{fraser1980conversational}.
Hence, we may naturally expect certain politeness strategies to signal that a conversation is likely to stay on track, while others might signal derailment. 

In particular, we consider a set of pragmatic devices 
signaling
 politeness drawn from \newcite{brown1987politeness}. These linguistic features reflect two overarching types of politeness. \textit{Positive} politeness strategies encourage social connection and rapport, perhaps serving to maintain cohesion throughout a conversation; such strategies include gratitude (``\textit{thanks} for your help''), 
 greetings (``\emph{hey}, how is your day so far'')
 and use of ``please'', both at the start (``\textit{Please} find sources for your edit...'') and in the middle (``Could you \textit{please} help with...?'')
 of a sentence. \textit{Negative} politeness strategies serve to dampen an interlocutor's imposition on an addressee, often through conveying indirectness or uncertainty on the part of the commenter. Both commenters in example \textbf{B} (Fig. \ref{fig:intro_examples}) employ one such strategy, hedging, perhaps seeking to soften an impending disagreement about a source's reliability (``I \textit{don't think}...'', ``I would \textit{assume}...'').
We also consider markers of {\em impolite} behavior, such as the use of direct questions (``\textit{Why}'s there no mention of it?') and sentence-initial 
second
 person pronouns (``\textit{Your} sources don't matter...''), which may serve as forceful-sounding contrasts to negative politeness markers.  Following \newcite{Danescu-Niculescu-Mizil2013-so}, we extract such strategies 
 by pattern matching on the dependency parses of comments.

\xhdr{Types of conversation prompts}
To complement our pre-defined set of politeness strategies, we seek to capture domain-specific rhetorical patterns used to initiate conversations.  For instance, in a collaborative setting, we may expect conversations that start with an invitation for working together to signal less tension between the participants than those that start with statements of dispute. 
We discover types of such \textit{conversation prompts} in an unsupervised fashion by  extending a framework used to infer the rhetorical role of questions in (offline) political debates \cite{Zhang+al:17c} to more generally extract the rhetorical functions of comments. 
The procedure follows the intuition that the rhetorical role of a comment is reflected in the type of replies it 
is likely to elicit.
  As such, comments which tend to trigger similar replies constitute a particular type of prompt.

To implement this intuition, we derive two different low-rank representations of the common lexical phrasings contained in comments (agnostic to the particular topical content discussed), automatically extracted as recurring sets of arcs in the dependency parses of comments. First, we derive \textit{reply-vectors} of phrasings, which reflect their propensities to \textit{co-occur}. In particular, we perform singular value decomposition on a term-document matrix $\mathcal{R}$ of phrasings and replies as $\mathcal{R} \approx \hat{\mathcal{R}} = U_R S V_R^T$, where rows of $U_R$ are low-rank reply-vectors for each phrasing.  

Next, we derive \textit{prompt-vectors} for the phrasings, which reflect similarities in the subsequent replies that a phrasing \textit{prompts}. 
We construct a prompt-reply matrix $\mathcal{P} = (p_{ij})$ where $p_{ij}=1$ if phrasing $j$ occurred in a reply to a comment containing phrasing $i$. We project $\mathcal{P}$ into the same space as $U_R$ by solving for $\hat{\mathcal{P}}$ in $\mathcal{P} = \hat{\mathcal{P}}SV_R^T$ as $\hat{\mathcal{P}} = \mathcal{P}V_R S^{-1}$. 
Each row of $\hat{\mathcal{P}}$ is then a prompt-vector of a phrasing, such that the prompt-vector for phrasing $i$ is close to the reply-vector for phrasing $j$ if comments with phrasing $i$ tend to prompt replies with phrasing $j$.
Clustering the rows of $\hat{\mathcal{P}}$ then yields $k$ conversational {\em prompt types} that are unified by their similarity in the space of replies. To infer the prompt type of a new comment, we represent the comment as an average of the representations of its constituent phrasings (i.e., rows of $\hat{\mathcal{P}}$) and assign the resultant vector to a cluster.\footnote{We scale rows of $U_R$ and $\hat{\mathcal{P}}$ to unit norm. We assign comments whose vector representation has ($\ell_2$)
 distance $\geq$ 1 to all cluster centroids to an extra, infrequently-occurring null type which we ignore in subsequent analyses.} 

To determine the prompt types of comments in our dataset, we first apply the above procedure to derive a set of prompt types from a {\em disjoint} 
(unlabeled)
 corpus of Wikipedia talk page conversations \cite{danescu2012echoes}. After initial examination of the framework's output on this 
 external data, 
  we chose to extract $k=6$ prompt types, shown in Table~\ref{tab:prompt_types} along with our interpretations.\footnote{We experimented with more prompt types as well, finding that while the methodology recovered finer-grained types, and obtained qualitatively similar results and prediction accuracies as described in Sections \ref{sec:analysis} and \ref{sec:prediction}, the assignment of comments to types was relatively sparse due to the small data size, resulting in a loss of statistical power.}  These prompts represent signatures of conversation-starters spanning a wide range of topics and contexts which reflect core elements of Wikipedia, such as 
  moderation
   disputes and coordination \cite{kittur2007he,Kittur2008-bc}.
We assign each comment in our present dataset to one of these types.\footnote{While the particular prompt types we discover are specific to Wikipedia, the methodology for inferring them is unsupervised and is applicable in other conversational settings.}

\section{Analysis}
\label{sec:analysis}
\begin{figure*}[ht]
\centering
\includegraphics[width=1\textwidth]{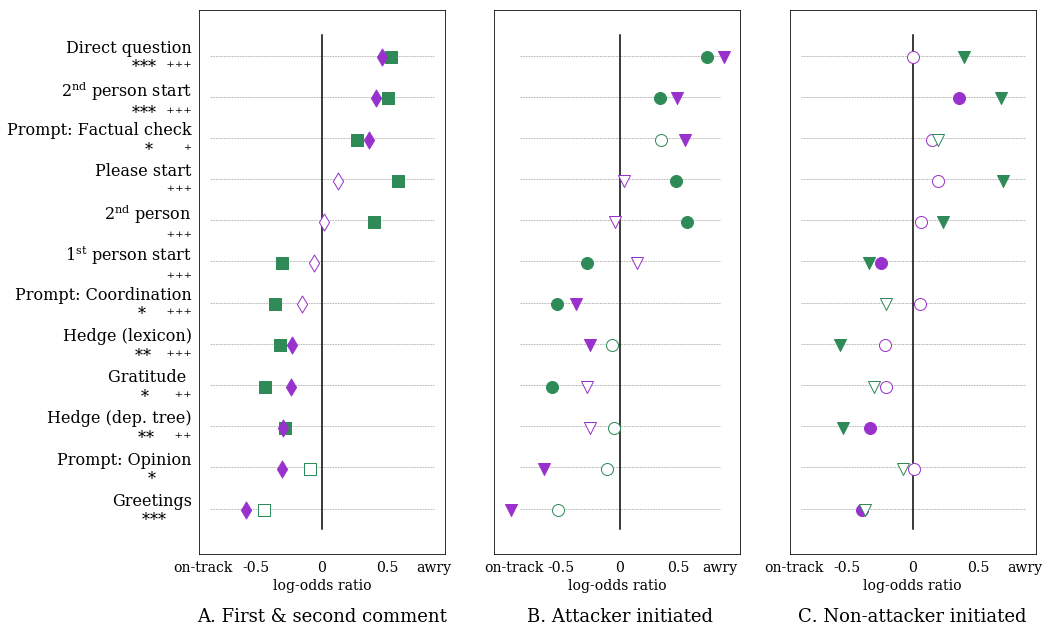}
\caption{
Log-odds ratios of politeness strategies and prompt types exhibited in the first and second comments of conversations that turn awry, versus those that stay on-track. \textbf{All}: {\textcolor{darkorchid}{Purple}} and {\textcolor{seagreen}{green}} markers denote log-odds ratios in the first and second comments, respectively; points are solid if they 
reflect significant (\mbox{$p < 0.05$}) log-odds ratios with an effect size of at least 0.2.
\textbf{A}: \textcolor{darkorchid}{$\diamondsuit$}s and \textcolor{seagreen}{$\square$}s denote \textbf{first} and \textbf{second} comment log-odds ratios, respectively; * denotes statistically significant differences at the \mbox{$p < 0.05$} (*), \mbox{$p < 0.01$} (**) and \mbox{$p < 0.001$} (***) levels for the first comment (two-tailed binomial test); + denotes corresponding statistical significance for the second comment.  \textbf{B} and \textbf{C}: $\triangledown$s and $\bigcirc$s correspond to effect sizes in the comments authored by the \textbf{attacker} and \textbf{non-attacker}, respectively, in \textbf{attacker initiated} (\textbf{B}) and \textbf{non-attacker initiated} (\textbf{C}) conversations.
}
\label{fig:feature_log_odds}
\end{figure*}

We are now equipped to computationally explore 
how the pragmatic devices used to start a conversation can signal its future health. Concretely, to quantify the relative propensity of a linguistic marker to occur at the start of awry-turning versus on-track conversations, we compute the log-odds ratio of the marker occurring in the initial exchange---i.e., in the first or second comments---of awry-turning conversations, compared to initial exchanges in the on-track setting. These quantities are depicted in Figure \ref{fig:feature_log_odds}A.\footnote{To reduce clutter we only depict features which occur a minimum of 50 times and have absolute log-odds $\geq 0.2$ in at least one of the data subsets.  The markers indicated as statistically significant for Figure \ref{fig:feature_log_odds}A remain so after a Bonferroni correction, with the exception of factual checks, hedges (lexicon, 
\textcolor{darkorchid}{$\diamondsuit$}), 
gratitude 
(\textcolor{darkorchid}{$\diamondsuit$}),
 and opinion.
}
Focusing on the \textbf{first} comment (represented as~\textcolor{darkorchid}{$\diamondsuit$}s), we find a rough correspondence between linguistic \textit{directness} and the likelihood of future personal attacks. In particular, comments which contain \textit{direct questions}, or exhibit \textit{sentence-initial you} (i.e., ``2$^{\text{nd}}$ person start''), tend to start awry-turning conversations significantly more often than ones that stay on track (both \mbox{$p< 0.001$}).\footnote{All $p$ values in this section are computed as two-tailed binomial tests,  comparing the proportion of awry-turning conversations exhibiting a particular device to the proportion of on-track conversations.} This effect coheres with our intuition that directness signals some latent hostility from the conversation's initiator, and perhaps reinforces the forcefulness of contentious impositions \cite{brown1987politeness}. 
This interpretation is also suggested by the relative propensity of the \texttt{factual check} prompt, which tends to cue disputes regarding an article's factual content ($p < 0.05$). 

In contrast, comments which initiate on-track conversations tend to contain \textit{gratitude} (\mbox{$p < 0.05$}) and \textit{greetings} \mbox{($p < 0.001$)}, both positive politeness strategies. Such conversations are also more likely to begin with \texttt{coordination} prompts \mbox{($p < 0.05$)}, signaling active efforts to foster constructive teamwork.  Negative politeness strategies are salient in on-track conversations as well, reflected 
by the use of
 \textit{hedges} (\mbox{$p < 0.01$}) and \texttt{opinion} prompts (\mbox{$p < 0.05$}), which may serve to soften impositions or factual contentions~\cite{hubler_understatements_1983}. 

These effects are echoed in the \textbf{second} comment---i.e.,
the \textbf{first reply} (represented as~\textcolor{seagreen}{$\square$}s). Interestingly, in this case we note that the difference in pronoun use is especially marked. First replies in conversations that eventually derail tend to contain more \textit{second person pronouns} (\mbox{$p<0.001$}), perhaps signifying a replier pushing back to contest the initiator; in contrast, on-track conversations have more 
\textit{sentence-initial~I/We}
 (i.e., ``1$^{\text{st}}$ person start'', \mbox{$p<0.001$}), potentially indicating the replier's willingness to step into the conversation and work with---rather than argue against---the initiator~\cite{tausczik2010psychological}.

\xhdr{Distinguishing interlocutor behaviors} Are the linguistic signals we observe solely driven by the eventual attacker, or do they reflect 
the behavior of both actors? To disentangle the attacker and non-attackers' roles in the initial exchange, we examine their language use in these two possible cases: when the 
\emph{future}
 attacker initiates the conversation, or is the first to reply. In \textbf{attacker-initiated} conversations (Figure \ref{fig:feature_log_odds}B, 608 conversations), we see that both actors exhibit a propensity for the linguistically direct markers (e.g., \textit{direct questions}) that tend to signal future attacks. Some of these markers are used particularly often by the \textbf{non-attacking replier} in awry-turning conversations (e.g., \textit{second person pronouns}, \mbox{$p < 0.001$}, \textcolor{seagreen}{$\bigcirc$}s), further suggesting the dynamic of the replier pushing back at---and perhaps even escalating---the attacker's initial hint of aggression. Among conversations initiated instead by the \textbf{non-attacker} (Figure \ref{fig:feature_log_odds}C, 662 conversations), the non-attacker's linguistic behavior in the first comment (\textcolor{darkorchid}{$\bigcirc$}s) is less distinctive from that of initiators in the on-track setting (i.e., log-odds ratios closer to 0); markers of future derailment are (unsurprisingly) more pronounced once the eventual attacker (\textcolor{seagreen}{$\triangledown$}s) joins the conversation in the second comment.\footnote{As an interesting avenue for future work, we note that 
some markers used by
 non-attacking initiators potentially still 
 anticipate 
later attacks, suggested by, e.g., the relative prevalence of \textit{sentence-initial you} (\mbox{$p < 0.05$}, \textcolor{darkorchid}{$\bigcirc$}s).}

More broadly, these results reveal how different politeness strategies and rhetorical prompts deployed in the initial stages of a conversation are tied to its future trajectory.

\section{Predicting Future Attacks}
\label{sec:prediction}
We now show that it is indeed feasible to predict whether a conversation will turn awry based on linguistic properties of its very first exchange, providing several baselines for this new task.  In doing so, we demonstrate that the 
pragmatic devices examined above encode signals about the future trajectory of conversations, capturing some of the intuition humans are shown to have.

We consider the following balanced prediction task: given a pair of conversations, which one will eventually lead to a personal attack? We extract all features from the very first exchange in a conversation---i.e., a comment-reply pair, like those illustrated 
in our introductory example (Figure~\ref{fig:intro_examples}).
We use logistic regression 
and report accuracies on a leave-one-page-out cross validation, 
such that in each fold, all conversation pairs from a given 
talk
 page are held out as test data and pairs from 
all
other pages are used as training data 
(thus preventing the use of page-specific information). Prediction results are summarized in Table~\ref{tab:prediction_results}.
 
\noindent\xhdr{Language baselines} As baselines, we consider several straightforward features: word count (which performs at chance level), sentiment lexicon \cite{liu2005opinion} and bag of words.
\noindent\xhdr{Pragmatic features} Next, we test the predictive power of the \textbf{prompt types} and \textbf{politeness strategies} features 
introduced in Section \ref{sec:features}. The 12 prompt type features (6 features for each comment in the initial exchange) achieve 59.2\% accuracy, and the 38 politeness strategies features (19 per comment) achieve 60.5\% accuracy.
The \textbf{pragmatic} features combine to reach 61.6\% accuracy. 

\noindent\xhdr{Reference points} To better contextualize the performance of our features, we compare their predictive accuracy to the following reference points:

\noindent\emph{Interlocutor features:} 
Certain kinds of interlocutors are potentially more likely to be involved in awry-turning conversations. For example, perhaps newcomers or anonymous participants are more likely to derail interactions than more experienced editors. We consider a set of features representing participants' experience on Wikipedia (i.e., number of edits) and whether the comment authors are anonymous. In our task, these features perform at the level of random chance.
\newcommand{\sectionrule}{\addlinespace[1ex]}

\begin{table}[t]
\centering
\begin{tabular}{lrc}
  \toprule
 \textbf{Feature set} &  \textbf{\# features} &  \textbf{Accuracy}  \\
  \midrule
  Bag of words & 5,000 & 56.7\%    \\
 Sentiment lexicon & 4 & 55.4\%  \\
\sectionrule
  \textbf{Politeness strategies} & 38 & 60.5\% \\
      \textbf{Prompt types} & 12 & 59.2\% \\
  \textbf{Pragmatic (all)} & 50 & 61.6\% \\
  \sectionrule
  \emph{Interlocutor features} & 5 & 51.2\% \\
  \emph{Trained toxicity} & 2 & 60.5\% \\
  \emph{Toxicity + \textbf{Pragmatic}} & 52 & 64.9\% \\
  \emph{Humans} & & 72.0\%  \\
  \bottomrule
\end{tabular}
\caption{Accuracies for the balanced future-prediction task. Features based on pragmatic devices are {\textbf{bolded}}, reference points are \emph{italicized}.
}
\label{tab:prediction_results}
\end{table}

\noindent\emph{Trained toxicity:} We also compare with the toxicity score of the exchange from the Perspective API classifier---a 
perhaps unfair 
reference point, since this supervised system was trained on additional human-labeled training examples from the same 
domain
 and since it was used to create the very 
data
 on which we evaluate. This results in an accuracy of 60.5\%; 
 combining trained toxicity with our
 pragmatic features achieves 64.9\%.

\noindent\emph{Humans:} A sample of 100 pairs 
  were labeled by (non-author) volunteer human annotators.  They were asked to guess, from the initial exchange, which conversation in a pair 
will
  lead to a personal attack.  Majority vote across three annotators was used to determine the human labels, resulting in an accuracy of 72\%. 
  This confirms that humans have some intuition about whether a conversation might be heading in a bad direction, which our features can partially capture.  In fact, the classifier 
  using pragmatic features is accurate on 80\% of the examples that humans also got right.

\noindent\xhdr{Attacks on the horizon} 
Finally, we seek to understand whether cues extracted from the first exchange can predict future discussion trajectory beyond the immediate next couple of comments.  We thus 
repeat the prediction experiments 
on the subset of conversations in which the first personal attack happens after the fourth comment (282 pairs), and find that the pragmatic devices used in the first exchange maintain their predictive power (67.4\% accuracy), 
while
the sentiment and bag of words baselines drop to the level of random chance.

Overall, these initial results show 
the feasibility of reconstructing some of the human intuition about the future trajectory of an ostensibly civil conversation 
in order to predict whether it will eventually turn awry.

\section{Conclusions and Future Work}
\label{sec:discussion}
In this work, we started to examine the intriguing phenomenon of conversational derailment, 
studying
how the use of pragmatic and rhetorical devices relates to future conversational failure.
Our investigation 
centers
 on the particularly perplexing scenario in which one participant of a civil discussion later attacks another, and explores the new task of predicting whether an initially healthy conversation 
will derail into such an attack. 
 To this end, we develop a computational framework for analyzing how 
 general
 politeness strategies and domain-specific rhetorical prompts 
 deployed in
  the initial stages of a conversation are tied to its future trajectory.

Making use of machine learning and crowdsourcing tools, we formulate a tightly-controlled setting that enables us to meaningfully compare conversations that stay on track with those that go awry. The human accuracy on predicting future attacks in this setting (72\%) suggests 
it is feasible at least at the level of human intuition.
 We show that our computational framework can recover some of that intuition, hinting at the potential of automated methods to identify signals of the future trajectories of online conversations.

Our approach has several limitations which open avenues for future work. Our correlational analyses do not provide any insights into \textit{causal} mechanisms of derailment, 
which randomized experiments could address. 
Additionally, 
since our procedure for collecting and vetting data focused on precision rather than recall, it might
miss
   more subtle attacks that are overlooked by the toxicity classifier. 
   Supplementing our investigation with other indicators of antisocial behavior, such as editors blocking one another, could enrich the range of attacks we study.  
Noting that our framework 
is not specifically tied to Wikipedia, it would also be valuable to 
explore the varied ways in which this phenomenon arises
in other (possibly non-collaborative) public discussion venues, such as Reddit and Facebook Pages.

While our analysis focused on the very first exchange in a conversation for the sake of generality, more complex modeling could extend 
its scope
to account for conversational features that more comprehensively span the interaction. Beyond the present binary classification task, one could explore a sequential formulation predicting whether the next turn is likely to be an attack as a discussion unfolds, capturing conversational dynamics such as sustained escalation.

Finally, our study of derailment offers only one glimpse into the space of possible conversational trajectories. Indeed, a manual investigation of conversations whose eventual trajectories were misclassified by our models---as well as 
by
the human annotators---suggests that interactions which initially seem prone to attacks can nonetheless maintain civility, by way of level-headed interlocutors, as well as explicit acts of reparation. A promising line of future work could consider the complementary problem of identifying pragmatic strategies that can help bring uncivil conversations back on track.

\vspace{0.15in}
\noindent\textbf{Acknowledgements.} We are grateful to the anonymous reviewers for their thoughtful comments and suggestions, and to 
Maria Antoniak,
Valts Blukis,
Liye Fu,
Sam Havron,
Jack Hessel,
Ishaan Jhaveri,
Lillian Lee,
Alex Niculescu-Mizil,
Alexandra Schofield,
Laure Thompson,
Andrew Wang,
Leila Zia and the members of the Wikimedia Foundation anti-harassment program
for extremely insightful (on-track) conversations and for assisting  with data annotation efforts.
This work is supported in part by  
NSF CAREER Award IIS-1750615,
 NSF Grant SES-1741441, 
a Google Faculty Award,
a 
WMF
gift 
and a 
CrowdFlower AI for Everyone Award.

\newpage

\bibliographystyle{acl_natbib}
\bibliography{paper-wiki}

\clearpage

\appendix

\section*{Appendix}

\section{Details on annotation procedure}
\label{sec:appendix_annot}

The process of constructing a labeled dataset 
for
 personal attacks 
 was challenging due to the complex and subjective nature of the 
 phenomenon,
  and developed over several iterations as a result. In order to guide future work, here we provide a detailed explanation of this process, expanding on the description in Section~\ref{sec:data}.

Our goal in this work was to understand linguistic markers of conversations that go awry and devolve into personal attacks---a highly subjective phenomenon with a multitude of possible definitions.\footnote{Refer to \citet{turnbull_thats_2018} for examples of challenges community moderators face in delineating personal attacks.} 
To enable a concrete analysis of conversational derailment that encompasses the scale and diversity of a setting like Wikipedia talk pages, we therefore needed to develop a well-defined conceptualization of conversational failure, and a procedure to accurately discover instances of 
this phenomenon
at scale. 

Our approach started from an {initial qualitative investigation} that resulted in a seed set of example conversational failures.  This seed set then informed the design of the subsequent {crowdsourced filtering} procedure, which we used to construct our full dataset.

\subsection{Initial qualitative investigation}
To develop our task, we compiled an initial sample of potentially awry-turning conversations by applying the candidate selection procedure (detailed in Section \ref{sec:data}) to a random subset of Wikipedia talk pages. This procedure yielded a set of conversations which the underlying trained classifier deemed to be initially civil, but with a later toxic comment. An informal inspection of these candidate conversations suggested many possible forms of toxic behavior, ranging from personal attacks (`Are you that big of a coward?'), to uncivil disagreements (`Read the previous discussions before bringing up this stupid suggestion again.'), to generalized attacks (`Another left wing inquisition?') and even to outright vandalism (`Wikipedia SUCKS!') or simply unnecessary use of foul language. 

Through our manual inspection, we also identified a few salient points of divergence between the classifier and our (human) judgment of toxicity. In particular, several comments which were machine-labeled as toxic were clearly sarcastic or self-deprecating, perhaps employing seemingly aggressive or foul language to bolster the collegial nature of the interaction rather than to undermine it. These false positive instances highlight the necessity of the subsequent crowdsourced vetting process---and point to opportunities to enrich the subtle linguistic and interactional cues such classifiers can address. 

\xhdr{Seed set} Our initial exploration of the automatically discovered candidate conversations and our discussions with the members of the Wikimedia Foundation anti-harassment program
pointed to a particularly salient and perplexing form of toxic behavior around which we centered our subsequent investigation: personal attacks \textit{from within}, where one of the two participants of the ostensibly civil initial exchange turns on another interlocutor. For each conversation where the author of the toxic-labeled comment also wrote the first or second comment, the authors manually checked that the interaction {started civil} and {ended in a personal attack}. The combined automatic and manual filtering process resulted in our {seed set} of 232 awry-turning conversations. 

We additionally used the candidate selection procedure to obtain on-track counterparts to each conversation in the seed set that took place on the same talk-page; this pairing protocol is further detailed in Section~\ref{sec:data}.

\xhdr{Human performance}
We gaged the feasibility of our task of predicting future personal attacks by asking (non-author) volunteer human annotators to label a 100-pair subset of the seed set. In this informal setting, also described in Section \ref{sec:prediction}, we asked each annotator to guess which conversation in a pair will lead to a personal attack on the basis of the initial exchange. Taking the majority vote across three annotators, the human guesses achieved an accuracy of 72\%, demonstrating that humans indeed have some systematic intuition for a conversation's potential for derailment.

\xhdr{Informing the crowdsourcing procedure}
To scale beyond the initial sample, we sought to use crowdworkers to replicate
 our process of manually filtering automatically-discovered candidates, enabling us to vet machine-labeled awry-turning and on-track conversations across the entire dataset. Starting from our seed set, we adopted an iterative approach to formulate our crowdsourcing tasks.

In particular, we designed an initial set of task instructions---along with 
definitions and examples of personal attacks---based on our 
observations of the seed set. Additionally, we chose a subset of conversations from the seed set to use as \textit{test questions} that crowdworker judgements on the presence or absence of such behaviors could be compared against. These test questions served 
both
as anchors to ensure the clarity of our instructions, 
and as quality controls. 
Mismatches between crowdworker responses and our own labels in trial runs then motivated subsequent modifications we made to the task design. The crowdsourcing jobs we ultimately used to compile our entire dataset are detailed below.

\subsection{Crowdsourced filtering} 
Based on our experiences in constructing and examining the seed set, we 
designed a crowdsourcing procedure to construct a larger set of personal attacks.
Here we provide more details about 
the crowdsourcing tasks, outlined in Section~\ref{sec:data}. We split the crowdsourcing procedure into two jobs, mirroring the manual process used to construct the seed set outlined above. The first job selected conversations ending with personal attacks; the second job enforced that \awryturning conversations start civil, and that \ontrack conversations remain civil throughout. We used the CrowdFlower platform to implement and deploy these jobs.
\xhdr{Job 1: Ends in personal attack} 
The first crowdsourcing job
was designed to select conversations containing a personal attack. 
In the annotation interface, each of three annotators was shown a candidate \awryturning conversation (selected using the procedure described in Section~\ref{sec:data}). The suspected toxic comment was highlighted, and workers were asked whether the highlighted comment contains a personal attack---defined in the instructions as a comment that is ``rude, insulting, or disrespectful towards a person/group or towards that person/group's actions, comments, or work.'' We instructed the annotators not to confuse personal attacks with civil disagreement, providing examples 
that illustrated this distinction.

To control the quality of the annotators and their responses,
we selected 82 conversations from the seed set to use as \textit{test questions}
with a known label.
Half of these test questions 
contained a personal attack and the other half were known to be 
civil.
The CrowdFlower platform's quality control tools automatically blocked workers who missed 
at least 20\% of these test questions.

While our task sought to identify personal attacks towards other interlocutors, trial runs of Job~1 suggested that many annotators construed attacks directed at other targets---such as groups or the Wikipedia platform in general---as personal attacks as well. To clarify the distinction between attack targets, and focus the annotators on labeling personal attacks, we asked annotators to specify \textit{who} the target of the attack is: (a) someone else in the conversation, (b) someone outside the conversation, (c) a group, or (d) other.
The resultant responses allowed us to filter annotations based on the reported target. This question also 
played the secondary role of ensuring that annotators read the entire conversation and accounted for this additional context in their choice.
In order to calibrate annotator judgements of what constituted an attack, we enforced that annotators saw a reasonable balance of awry-turning and on-track conversations. By virtue of the candidate selection procedure, a large proportion of the conversations in the candidate set contained attacks. Hence, we also included 804 candidate on-track conversations in the task.

Using the output of Job 1, we filtered our candidate set to the conversations where \textit{all three annotations} agreed that a personal attack had occurred. We found that unanimity produced higher quality labels than taking a majority vote 
by omitting ambiguous cases (e.g., the comment ``It's our job to document things that have received attention, however ridiculous we find them.'' could be insulting towards the things being documented, but could also be read as a statement of policy).\footnote{This choice further sacrifices recall for the sake of label precision, an issue that is also discussed in Section \ref{sec:discussion}.}

\xhdr{Job 2: Civil start} 
The second crowdsourcing job 
was designed to enforce that candidate \awryturning conversations start civil, and candidate \ontrack conversations remain civil throughout. 
Each of three annotators was shown comments from both \ontrack and \awryturning conversations that had already been filtered through Job 1. They were asked whether any of the displayed comments were toxic---defined as ``a rude, insulting, or disrespectful comment that is likely to make someone leave a discussion, engage in fights, or give up on sharing their perspective.'' This definition was adapted from previous efforts to annotate toxic behavior \cite{toxicityDataset2017} and intentionally targets a broader spectrum of uncivil behavior.  

 As in Job 1, we instructed annotators to not confound civil disagreement with toxicity. To reinforce this distinction, we included an additional question asking them whether any of the comments displayed disagreement, and prompted them to identify particular comments.
Since toxicity can be context-dependent, we wanted annotators to have access to the full conversation to help inform their judgement about each comment. 
However, we were also concerned that 
annotators would be overwhelmed by the amount of text in long conversations, 
and might be deterred from carefully reading each comment as a result. Indeed, in a trial run where full conversations were shown, we received negative feedback from annotators regarding task difficulty. 
To mitigate this difficulty without entirely omitting contextual information, we 
divided each conversation into snippets of three comments each. This 
kept
 the task fairly readable while 
still providing
some local context. For candidate \awryturning conversations, we generated the snippets from all comments except the last one (which is known from Job 1 to be an attack).
For \ontrack conversations, we generated the snippets from all comments in the conversation.

We marked conversations as toxic if at least three annotators, across all snippets of the conversation, identified at least one toxic comment. As in Job 1, we found that requiring this level of consensus among annotators produced reasonably high-quality labels.
\noindent\xhdr{Overall flow} 
To compile our full dataset, we started 
with 3,218 candidate \awryturning conversations which were filtered using Job 1, 
and discarded all but 435 conversations which all three annotators 
labeled as ending in a personal attack towards someone else in the conversation. 
These 435 conversations, along with paired \ontrack conversations, were then filtered using Job 2. 
This step removed 30 pairs: 24 where the \awryturning conversation was found to contain toxicity before the personal attack happened, and 6 where the \ontrack conversation was found to contain toxicity. We combined the crowdsourced output with the seed set to obtain a final dataset of 1,270 paired \awryturning and \ontrack conversations.

\section{Further examples of prompt types}
\label{sec:prompt_type_egs}

\begin{table*}[t]
\centering
\begin{adjustbox}{width=2.1\columnwidth}
  \begin{tabular}{lll}
    \toprule
    Prompt Type & Example comments & Example replies \\
    \midrule
    \texttt{Factual check} & I \textbf{don't see} how this \textbf{is} relevant. & I \textbf{don't understand} your dispute. \\
    & \textbf{Therefore} 10 knots \textbf{is} 18.5 miles per hour. &  This \textbf{means} he \textbf{is} unlikely to qualify as an expert. \\
    & That \textbf{does not mean} you can use this abbreviation everywhere. &  They did \textbf{not believe} he will return. \\
    & ``Techniques'' \textbf{refer} specifically to his fighting. &  I \textbf{disagree}. \\
    \addlinespace[5pt]
    \texttt{Moderation} & \textbf{Please stop} making POV edits to the article. & I\textbf{'ve reverted} your change [...] \\
    & Your edits appear to be vandalism and have \textbf{been reverted}. & I\textbf{'ve asked} them to stop. \\
    & I \textbf{have removed} edits which seem nationalistic. & The next occurrance will \textbf{result in} a block. \\
    & These mistakes should \textbf{not} be \textbf{allowed} to remain in the article. & \textbf{Do not remove} my question. \\
    \addlinespace[5pt]
    \texttt{Coordination}  & I \textbf{have been working on} creating an article.  & If you can do it I \textbf{would appreciate it}. \\
    & \textbf{Feel free} to correct my mistake. & I \textbf{have to go} but I'll be back later. \\
    & I \textbf{expanded} the article from a stub. & \textbf{Ok, thanks}.\\
    & I\textbf{'ll make sure} to include a plot summary. & \textbf{Hopefully} it will \textbf{be} fixed in a week.\\
    \addlinespace[5pt]
    \texttt{Casual remark} & \textbf{Just to} save you any issue in the future [...] & \textbf{Yeah}, this has \textbf{gotten} out of hand. \\
    & \textbf{Remember} that badge I gave you? & \textbf{Anyway}, it\textbf{'s nice} to see you took the time [...] \\
    & \textbf{Oh}, that\textbf{'s} fabulous, \textbf{love} the poem! & \textbf{Yep}, that\textbf{'s cool}. \\
    & \textbf{Not sure} how that last revert came in there. & I \textbf{just thought} your comment was no longer needed.\\
    \addlinespace[5pt]
    \texttt{Action statement}  & \textbf{If} you have \textbf{uploaded} other media, \textbf{consider} checking the criteria.  & That article has been \textbf{tagged for} deletion. \\
    & \textbf{Could} somebody \textbf{please explain} how they differ?  & I\textbf{'ve fixed} the wording. \\
    & The info \textbf{was placed} in the appropriate section.  & \textbf{Replaced with} free picture for all pages. \\
    & Could you \textbf{undelete} my article?  & It \textbf{has been deleted by} an admin. \\
    \addlinespace[5pt]
    \texttt{Opinion}  & I've been \textbf{thinking of} setting up a portal.  & It \textbf{seems} very much \textbf{in} the Wiki spirit. \\
    & I \textbf{am wondering} if he is not supposed to be editing here. & \textbf{Sounds like} a good idea. \\
    & It\textbf{'s hard} to combine these disputes. & I \textbf{also think} we \textbf{need} to clarify this. \\
    \bottomrule
  \end{tabular}
\end{adjustbox}
\caption{Further examples of representative comments in the data for each automatically-extracted prompt type, along with examples of typical replies prompted by each type, produced by the methodology outlined in Section~\ref{sec:features}. 
Bolded text indicate common prompt and reply phrasings identified by the framework in the respective examples; 
note that the comment and reply examples in each row do not necessarily correspond to one another.}
\label{tab:prompt_types_appendix}
\end{table*}

Table \ref{tab:prompt_types_appendix} provides further examples of comments containing the prompt types we automatically extracted from talk page conversations using the 
unsupervised methodology described in Section \ref{sec:features}; descriptions of each type can be found in Table \ref{tab:prompt_types}. For additional interpretability, we also include examples of typical \textit{replies} to comments of each prompt type, which are also extracted by the method.

\end{document}